\documentclass[conference]{IEEEtran}
\usepackage[left=0.635in,right=0.635in,top=0.75in,bottom=1.05in]{geometry}

\usepackage[numbers]{natbib}
\usepackage[font=small,skip=0pt]{caption}
\usepackage{subcaption}
\usepackage{array,multirow,booktabs}

\usepackage{tabularray}
\usepackage{algorithmic}
\usepackage{graphicx}
\usepackage{textcomp}
\usepackage{fancyhdr}
\usepackage[table]{xcolor} 
\usepackage{colortbl}
\usepackage{makecell}
\usepackage{tabularx}
\usepackage{float}
\usepackage{amsmath}
\usepackage{amssymb}
\usepackage{amsfonts}
\usepackage{listings}
\usepackage{xcolor}
\usepackage[most]{tcolorbox}
\usepackage{subcaption}
\usepackage{tikz}
\usepackage{caption}
\usepackage{enumitem}
\usepackage{hyperref}
\usepackage{subcaption} 
\definecolor{PineGreenDarkest}{HTML}{4C9E8F}
\definecolor{PineGreenDark}{HTML}{70B3A1}
\definecolor{PineGreenMedium}{HTML}{94C8B4}
\definecolor{PineGreenLight}{HTML}{B8DDC6}
\definecolor{PineGreenLighter}{HTML}{D1EED9}
\definecolor{darkgreen}{rgb}{0.0, 0.5, 0.0} 
\lstdefinestyle{mypython}{
    language=Python,
    basicstyle=\ttfamily\small,
    keywordstyle=\color{blue}\bfseries,
    stringstyle=\color{darkgreen},
    commentstyle=\color{gray}\itshape,
    morekeywords={BaseModel, Field, List},
    columns=flexible,
    keepspaces=true,
    breaklines=true,
    showstringspaces=false,
    frame=single,
    rulecolor=\color{black},
    stepnumber=1,
    numbersep=5pt,
    xleftmargin=0.05\columnwidth,   
    xrightmargin=0.05\columnwidth,  
}
\renewcommand{\arraystretch}{0.75}

\begin{document}
%
\title{Vision Language Models for Optimization-Driven Intent Processing in Autonomous Networks}

\author{
    \IEEEauthorblockN{Tasnim Ahmed, Yifan Zhu,
    Salimur Choudhury}
    
    \IEEEauthorblockA{School of Computing,
    Queen's University, Ontario, Canada}

    \IEEEauthorblockA{\{tasnim.ahmed, yifan.zhu, s.choudhury\}@queensu.ca}
}
\IEEEspecialpapernotice{Accepted for presentation at \href{https://icc2026.ieee-icc.org/}{The IEEE International Conference on Communications (ICC) 2026}}
\maketitle

\begin{abstract}
Intent-Based Networking (IBN) allows operators to specify high-level network goals rather than low-level configurations. 
While recent work demonstrates that large language models can automate configuration tasks, a distinct class of intents requires generating optimization code to compute provably optimal solutions for traffic engineering, routing, and resource allocation. 
Current systems assume text-based intent expression, requiring operators to enumerate topologies and parameters in prose. 
Network practitioners naturally reason about structure through diagrams, yet whether Vision-Language Models (VLMs) can process annotated network sketches into correct optimization code remains unexplored.
We present IntentOpt, a benchmark of 85 optimization problems across 17 categories, evaluating four VLMs (GPT-5-Mini, Claude-Haiku-4.5, Gemini-2.5-Flash, Llama-3.2-11B-Vision) under three prompting strategies on multimodal versus text-only inputs.
Our evaluation shows that visual parameter extraction reduces execution success by 12–21 percentage points (pp), with GPT-5-Mini dropping from 93\% to 72\%. Program-of-thought prompting decreases performance by up to 13 pp, and open-source models lag behind closed-source ones, with Llama-3.2-11B-Vision reaching 18\% compared to 75\% for GPT-5-Mini. These results establish baseline capabilities and limitations of current VLMs for optimization code generation within an IBN system.
We also demonstrate practical feasibility through a case study that deploys VLM-generated code to network testbed infrastructure using Model Context Protocol.
\end{abstract}


\IEEEpeerreviewmaketitle

\section{Introduction}
Network operators today configure hundreds of interdependent parameters across routing protocols, quality-of-service policies, and resource allocators. As networks incorporate virtualized functions, edge computing, and dynamic slicing, manual configuration has become a bottleneck. Experts spend days translating business requirements into vendor-specific commands, and every change risks introducing errors. Intent-Based Networking (IBN) offers a path forward by letting operators specify what the network should achieve rather than how to achieve it. Instead of writing router commands manually, an operator states ``provide 100 Mbps throughput for 10,000 mobile users,'' and the system determines the configuration. Recent work demonstrates that Large Language Models (LLMs) can automate parts of this translation process. These systems convert natural language goals into deployable artifacts across diverse management tasks. For example, researchers have used LLMs to generate service descriptors for 5G mobile cores, producing deployment templates from textual requirements \cite{dinh2025endtoend, brodimas2025intent5g}. Others target device configuration, translating intents into Border Gateway Protocol policies \cite{fuad2024intent}, firewall rules \cite{fuad2024intent}, and software-defined networking programs in P4 and eBPF \cite{angi2025llnet}. Production systems now orchestrate multi-step network workflows. Confucius framework from Meta processes over 240,000 sessions per month, chaining diagnosis and capacity planning tasks through validated tool invocations \cite{wang2025confucius}. Standards-aligned pipelines convert operator requests into TMForum service orders and deploy them through operational support systems \cite{trantzas2025intentdriven}. Hierarchical architectures combine intent processing, feasibility validation, and closed-loop execution for RANs \cite{habib2025generative}. At the earliest stage, intent extraction modules recognize management actions such as deployment, modification, and reporting from unstructured operator requests \cite{manias2024intentextraction}.
All of these advances have a shared feature in that they deal with intents whose requirements correspond directly to configuration parameters.
The system matches the request to a template, fills in values such as throughput targets or access rules, validates syntax, and deploys the result. However, a distinct class of intents requires an intermediate optimization step before configuration can be determined. Consider an operator who states ``minimize the cost of routing traffic between datacenters while meeting latency requirements.'' Achieving similar goals requires formulating a mathematical optimization problem, solving it, and translating the result into deployable configurations. Such intents arise in tasks like traffic engineering using minimum cost flow models and route planning using traveling salesman formulations, among many others. Unlike simpler cases, these tasks require generating and executing optimization code before producing configuration artifacts.
Existing LLM-based intent systems do not address this gap. The research works described above assume configuration parameters are either specified directly or derived through policy matching, but they do not synthesize optimization models. Optimization intents refer to a distinct class of intents where achieving the desired outcome requires solving an intermediate optimization problem, such as routing or resource matching, before the final configuration can be determined. The distinction matters because optimization intents couple topology understanding with algorithmic modeling. The system must interpret network structure, identify key parameters such as capacities and costs, formulate and execute the appropriate optimization model, and derive the resulting configuration. Errors in any of these stages, such as incorrect parameter extraction or constraint omission, can propagate to the final configuration and compromise service guarantees.

Current IBN systems assume operators express all information as text, listing node identifiers, link endpoints, capacities, and costs in prose. For a six-node network, this leads to verbose and error-prone descriptions such as ``Node 1 connects to node 2 with capacity 50 and cost 10.'' In practice, practitioners already reason visually, sketching topologies and annotating links with parameters. A natural extension is to allow annotated network diagrams as intent inputs. Vision-Language Models (VLMs) have shown strong performance on multimodal reasoning tasks, yet their ability to interpret network diagrams and generate correct optimization code remains untested. Such a model must extract topology from sketch, parse numeric labels, map the problem to an optimization family, and produce executable solver code.
Existing benchmarks do not evaluate this capability. Code generation benchmarks test algorithmic reasoning on text-only function specifications without multimodal inputs or optimization modeling \cite{humaneval}. Multimodal reasoning benchmarks ChartQA \cite{chartqa} assess numeric question answering over charts and geometry diagrams but do not require generating executable code or invoking external solvers. Existing resources do not combine topology understanding, visual parameter extraction, and optimization code generation in a single evaluation framework.

To this end, we present IntentOpt, a benchmark of 85 network optimization problems spanning 4 families and 17 categories (refer to Table \ref{tab:problem_taxonomy}). Each instance includes an annotated network diagram, a natural language intent, ground truth Gurobi code verified to produce optimal solutions, and a text-only variant to isolate visual reasoning effects. We evaluate four vision language models (GPT 5 Mini, Claude Haiku 4.5, Gemini 2.5 Flash, and Llama 3.2 11B Vision) using three prompting strategies, namely direct, role, and program of thought (PoT). Our evaluation reveals key insights into the capabilities and limitations of VLMs for multimodal optimization intent processing. Visual parameter extraction remains a major bottleneck compared to text inputs, PoT prompting often hinders code synthesis despite its reasoning strength, and open-source models still trail behind closed-source ones.

This work makes four contributions. (1) It introduces IntentOpt, a benchmark of 85 verified network optimization problems featuring multimodal inputs paired with ground truth solver code. (2) It presents the first systematic evaluation of vision language models on optimization code generation, quantifying the vision language gap and identifying key failure modes across model classes and prompting strategies. (3) It demonstrates practical feasibility through a case study that deploys VLM generated optimization code on a network testbed using Model Context Protocol (MCP). (4) It establishes a reproducible evaluation framework to support future research on multimodal intent understanding in autonomous networks.

\section{The IntentOpt Benchmark}
\label{sec:benchmark}

\begin{figure}[!t] 
  \centering
  \includegraphics[width=\columnwidth]{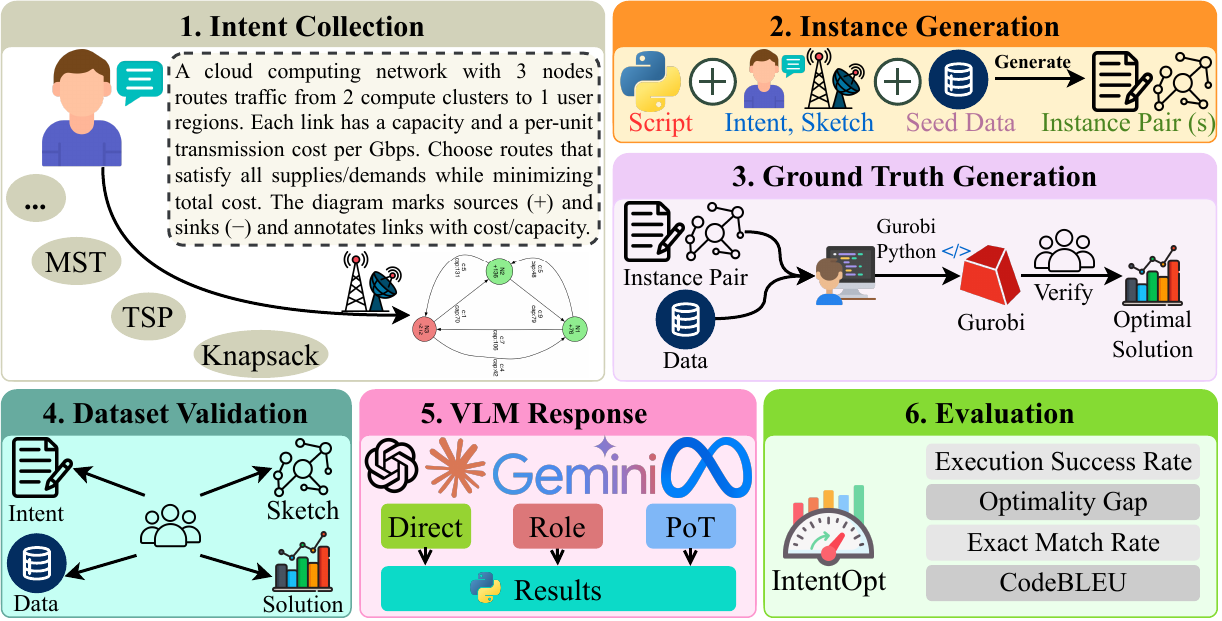}
  \caption{\textbf{Six stage pipeline for IntentOpt construction and evaluation.} Stages 1 to 4 create the benchmark and stages 5 and 6 evaluate VLMs using different prompting strategies.}
  \label{fig:intentopt-pipeline}
\end{figure}

We present IntentOpt, a benchmark for evaluating VLMs on network optimization code generation. Each instance includes a labeled network diagram, a natural language intent specifying objectives and constraints, and Gurobi code. To assess visual reasoning independently, a text-only variant provides all numeric parameters within the intent description. Our dataset construction comprises four stages: intent collection, instance generation, ground truth generation, and dataset validation. Figure~\ref{fig:intentopt-pipeline} illustrates the complete benchmarking framework. We have made the benchmark publicly available at \href{https://github.com/tasnim7ahmed/IntentOpt}{github.com/tasnim7ahmed/IntentOpt}.

\textbf{Stage 1: Intent Collection.} 
We curate 17 problem categories that represent key network management tasks across four families, namely Geometric and Combinatorial, Flow Conservation, Assignment and Transportation, and Bounded Flow. Tour-based optimization includes Traveling Salesman and Minimum Latency, backbone design uses Minimum Spanning Tree, and connectivity queries use Shortest Path. Traffic engineering includes Minimum Cost Flow, Capacitated Flow, Single and Multi Commodity Flow, Circulation, and Max Flow. Resource allocation includes Assignment, Transportation, and Knapsack, while facility placement is represented by K Center. Network Simplex, Transshipment, and Lower and Upper Bound Flow address cost routing, intermediate storage, and quality of service constraints. These problems are instantiated on bipartite, geometric, directed, and general weighted graphs, requiring models to handle diverse constraints such as flow conservation, capacity limits, degree bounds, and cardinality restrictions. Table~\ref{tab:problem_taxonomy} shows the taxonomy, applications, and sizes.


\begin{table}[t]
\scriptsize
\centering
\setlength{\tabcolsep}{2pt}
\renewcommand{\arraystretch}{0.98}
\caption{Problem taxonomy in the IntentOpt benchmark grouped by family. Each category includes 5 instances.}
\label{tab:problem_taxonomy}
\begin{tabular}{@{}p{1.75cm}p{2.55cm}p{2.35cm}p{0.95cm}@{}}
\toprule
\textbf{Family} & \textbf{Category} & \textbf{Application} & \textbf{Nodes} \\
\midrule
\multirow{5}{1.75cm}{\begin{tabular}[t]{@{}l@{}}Geometric\\and\\Combinatorial\end{tabular}}
 & Traveling Salesman & Route optimization & 5--10 \\
 & Minimum Latency & Delay sens.\ routing & 5--10 \\
 & Minimum Spanning Tree & Backbone design & 5--10 \\
 & Knapsack & Provisioning & N/A \\
 & K Center & Facility placement & 5--10 \\
\midrule
\multirow{6}{1.75cm}{\begin{tabular}[t]{@{}l@{}}Flow\\Conservation\end{tabular}}
 & Min Cost Flow & Traffic engineering & 2--6 \\
 & Capacitated Flow & Bandwidth alloc. & 2--6 \\
 & Single Commodity Flow & Single tenant routing & 2--6 \\
 & Multi Commodity Flow & Multi tenant routing & 2--6 \\
 & Circulation & Load balancing & 2--5 \\
 & Max Flow & Throughput max. & 2--5 \\
\midrule
\multirow{3}{1.75cm}{\begin{tabular}[t]{@{}l@{}}Assignment\\and\\Transportation\end{tabular}}
 & Assignment & Resource alloc. & 2--7 \\
 & Transportation & Demand satisfaction & 2--7 \\
 & Transshipment & Warehouse routing & 2--4 \\
\midrule
\multirow{3}{1.75cm}{\begin{tabular}[t]{@{}l@{}}Bounded\\Flow\end{tabular}}
 & Bounded Flow & QoS guarantees & 2--6 \\
 & Network Simplex & Flow with costs & 2--6 \\
 & Shortest Path & Path planning & 3--6 \\
\bottomrule
\end{tabular}
\end{table}

\textbf{Stage 2: Instance Generation.}
For each category, we generate five instances with varying complexity. We sample node counts, edge densities, capacities, costs, and demands from ranges calibrated to avoid trivial solutions while maintaining feasibility under standard solver time limits. Furthermore, we remove headers and filenames to prevent models from inferring problem type through metadata rather than diagram content. We provide two input variants for each instance. The text-only version offers a structured description with explicit numeric parameters such as node identifiers, edges, capacities, costs, and demands. The multimodal version combines a network diagram with a natural language intent describing objectives and constraints without numeric details, requiring models to extract values from visual labels. Both variants express objectives and constraints in natural language without exposing mathematical formulations or decision variables. Automated checks ensure parameter validity, feasibility, and consistency, and each instance is verified to be solvable within practical computational limits.

\textbf{Stage 3: Ground Truth Generation.}
We generate executable Gurobi Python code for each instance, covering parameter extraction, variable declaration, objective and constraint definition, and solver execution with result extraction. The Gurobi solver computes optimal solutions, and corresponding objective values are recorded. Instances with solver errors or numerical issues are flagged for review.

\textbf{Stage 4: Dataset Validation.}
Validation proceeds in three stages to ensure correctness. Automated tests first confirm that the code runs successfully and the solver returns an optimal solution. Human reviewers then check modeling accuracy, verifying that variables, objectives, and constraints align with the intended problem description. Finally, cross-validation is performed by solving selected instances with alternative formulations, ensuring objective values remain consistent within a 1\% tolerance. Instances showing inconsistency or numerical instability are excluded.

\section{Evaluation Framework}
\label{sec:methodology}

\subsection{Problem Formulation}
We define network optimization intent processing as a conditional code generation task. Each instance provides a network diagram, a natural language intent, and ground truth Gurobi code with a verified optimal objective value. Given the diagram and intent, a VLM generates executable Python code, which is executed to compare the obtained objective value with the reference solution. This setup evaluates two abilities: extracting numeric parameters from visual inputs and generating correct optimization formulations. The text-only variant isolates the latter by removing visual reasoning requirements.

\subsection{Models}
We evaluate four VLMs that represent both closed source and open source approaches, combining state-of-the-art performance with architectural and accessibility diversity. The closed-source models include GPT-5-Mini from OpenAI, Claude-Haiku-4.5 from Anthropic, and Gemini-2.5-Flash from Google. The open source model, Llama-3.2-11B Vision from Meta, serves as a publicly available alternative for assessing whether open models can approach closed source performance on optimization code generation tasks. Differences in model scale further provide insight into performance variation across architectures. All models are evaluated under identical settings using temperature zero for deterministic decoding.

\subsection{Prompting Strategies}

Prompt design significantly influences language model performance. We investigate three prompting strategies that represent different approaches to guiding model behavior.

\subsubsection{Direct Prompting} The model receives the network diagram, intent description, and a straightforward instruction to generate executable Gurobi code. This strategy provides minimal guidance, testing whether models can perform end-to-end reasoning from visual input to working code without additional scaffolding. The prompt specifies the expected output format and emphasizes the need for syntactically correct, complete implementations.

\subsubsection{Role Prompting} We augment the instruction by positioning the model as an expert in network optimization and mathematical programming. The prompt frames the task as one requiring domain expertise, stating that the model should approach the problem as a specialist would. This tests whether persona-based conditioning improves performance by activating relevant knowledge and reasoning patterns.

\subsubsection{Program-of-Thought} The model is asked to generate a code that includes explicit comments explaining its reasoning process. Comments should describe how the problem is decomposed, what each decision variable represents, and why each constraint is necessary. This approach encourages structured problem solving and makes the reasoning process transparent. Unlike chain-of-thought prompting that produces natural language reasoning before code, PoT integrates reasoning directly into code comments.

\subsection{Evaluation Metrics}

We evaluate model performance using 4 metrics that capture different aspects of code quality and correctness.

\subsubsection{Execution Success Rate (ESR)} This metric measures the percentage of instances where generated code runs without errors and produces a solution.
\subsubsection{Optimality Gap} For successfully executed instances, we measure solution quality by comparing the returned objective value with the ground truth. The optimality gap, defined as the absolute difference divided by the ground truth value and expressed as a percentage, represents the relative error. We report the median gap across all successful runs as a robust indicator of typical performance unaffected by large outliers.

\subsubsection{Exact Match Rate (EMR)}
This metric is defined as the proportion of instances whose optimality gap is below 1\%, accounting for minor numerical precision differences and representing solutions that are practically optimal.

\subsubsection{CodeBLEU}
CodeBLEU is used to measure structural similarity between generated and reference code independent of outcomes. It extends the standard BLEU metric by incorporating abstract syntax tree matching and data flow analysis.

\subsection{Statistical Analysis}
McNemar’s test is applied to binary metrics such as ESR and EMR, and the Wilcoxon signed rank test is used for continuous metrics such as the optimality gap.
Bootstrap confidence intervals with 10,000 resamples quantify estimation uncertainty. Effect sizes are calculated using Cohen’s d, where values below 0.2 indicate small, between 0.2 and 0.8 indicate medium, and above 0.8 indicate large practical significance.

\section{Results}
Table \ref{tab:main_results} summarizes overall performance across all evaluated models and prompting strategies on the full set of 85 problem instances for both multimodal (diagram + text) and text-only inputs. GPT-5-Mini achieves the highest execution success at 75.29\% under role prompting on multimodal inputs, with 0.0\% median optimality gap and 51.76\% exact match rate. Text-only performance reaches 91.76\%, giving a 16.47 pp vision language gap that ranges from 16.47 (role) to 25.89 (PoT). Claude-Haiku-4.5 and Gemini-2.5-Flash show similar trends, with gaps of 17.65 to 18.82 and 5.88 to 11.77 pp respectively. All closed-source models record 0.0\% median gaps on text-only inputs, confirming reliable optimization when parameters are extracted correctly. Llama-3.2-11B-Vision shows comparable execution success between multimodal (17.65\%) and text-only (16.47\%) settings for direct prompting, indicating that visual understanding is not its main limitation.

\begin{table*}[t]
\tiny
\centering
\caption{Performance comparison between multimodal and text only inputs across 85 problem instances.}
\label{tab:main_results}
\resizebox{\textwidth}{!}{%
\begin{tabular}{llcccccccc}
\toprule
& & \multicolumn{4}{c}{\textbf{Multimodal}} & \multicolumn{4}{c}{\textbf{Text Only}} \\
\cmidrule(lr){3-6} \cmidrule(lr){7-10}
\textbf{Model} & \textbf{Prompt} & \textbf{ESR (\%)} & \textbf{Gap (\%)} & \textbf{EMR (\%)} & \textbf{CodeBLEU} & \textbf{ESR (\%)} & \textbf{Gap (\%)} & \textbf{EMR (\%)} & \textbf{$\Delta$ ESR (pp)} \\
\midrule
\multirow{3}{*}{GPT-5-Mini} 
 & direct & 71.76 & 0.0 & 44.71 & 19.06 & 92.94 & 0.0 & 85.88 & 21.18 \\
 & role & 75.29 & 0.0 & 51.76 & 20.82 & 91.76 & 0.0 & 85.88 & 16.47 \\
 & PoT & 62.35 & 0.0 & 43.53 & 19.51 & 88.24 & 0.0 & 78.82 & 25.89 \\
\midrule
\multirow{3}{*}{Claude-Haiku-4.5}
 & direct & 62.35 & 12.81 & 29.41 & 23.32 & 80.0 & 0.0 & 65.88 & 17.65 \\
 & role & 60.0 & 4.57 & 30.59 & 23.82 & 78.82 & 0.0 & 69.41 & 18.82 \\
 & PoT & 50.59 & 12.53 & 22.35 & 22.67 & 69.41 & 0.0 & 58.82 & 18.82 \\
\midrule
\multirow{3}{*}{Gemini-2.5-Flash}
 & direct & 62.35 & 0.0 & 41.18 & 21.59 & 74.12 & 0.0 & 67.06 & 11.77 \\
 & role & 65.88 & 0.0 & 37.65 & 21.96 & 72.94 & 0.0 & 67.06 & 7.06 \\
 & PoT & 60.0 & 0.0 & 32.94 & 21.7 & 65.88 & 0.0 & 60.0 & 5.88 \\
\midrule
\multirow{3}{*}{Llama-3.2-11B}
 & direct & 17.65 & 100.0 & 1.18 & 16.27 & 16.47 & 89.18 & 2.35 & $-$1.18 \\
 & role & 7.06 & 88.87 & 0.0 & 16.98 & 12.94 & 0.0 & 7.06 & 5.88 \\
 & PoT & 1.18 & 0.0 & 1.18 & 14.83 & 12.94 & 66.11 & 4.71 & 11.76 \\
\bottomrule
\end{tabular}%
}
\end{table*}

\subsection{Effect of Prompting Strategy}

\begin{figure*}[t]
\centering
\includegraphics[width=0.9\textwidth]{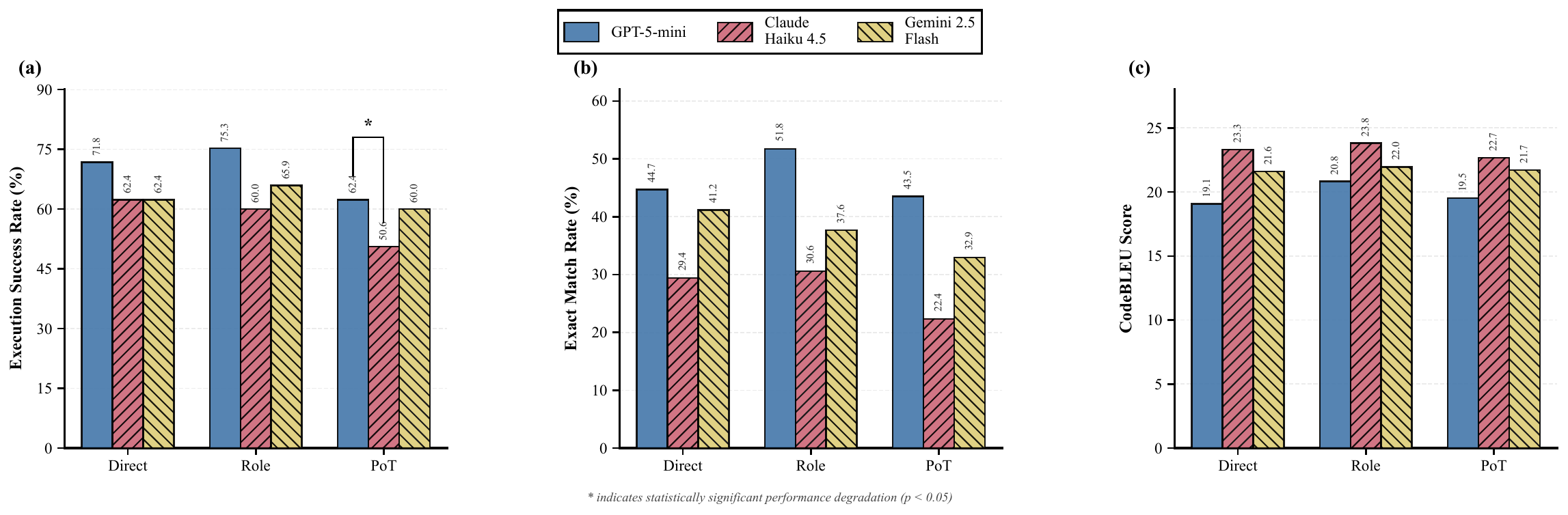}
\caption{\textbf{Effect of prompting strategy on model performance (multimodal).} (a)  PoT reduces ESR for GPT-5-Mini and Claude-Haiku-4.5 ($p < 0.05$). (b) EMR shows similar trends. (c) CodeBLEU remains stable, indicating structural quality is less impacted.}

\label{fig:prompting_strategies}
\end{figure*}

Figure~\ref{fig:prompting_strategies} compares prompting strategies across three metrics. For GPT-5-Mini, role prompting improves over direct (75.29\% versus 71.76\%) but not significantly (p = 0.6291), while PoT degrades to 62.35\% (p = 0.0266 versus role). Claude-Haiku-4.5 shows similar degradation under PoT (62.35\% to 50.59\%, p = 0.0414). Gemini-2.5-Flash shows no significant differences across strategies (all p $>$ 0.42). CodeBLEU scores remain stable across prompting strategies, indicating structural code quality is less affected than functional correctness. This degradation contrasts with mathematical reasoning benchmarks where explicit steps improve accuracy.

\subsection{Statistical Analysis}

McNemar's tests and bootstrap confidence intervals confirm statistical significance. For prompting strategies, GPT-5-Mini shows significance between role and PoT (p = 0.0266) but not for other pairs. Claude-Haiku-4.5 shows significance for direct versus PoT (p = 0.0414). Gemini-2.5-Flash shows no significant differences. Bootstrap 95\% confidence intervals for GPT-5-Mini multimodal execution success are [62.4, 81.2] (direct), [65.9, 84.7] (role), and [51.8, 72.9] (PoT). For Llama-3.2-11B-Vision, intervals are [9.4, 25.9] (direct), [2.4, 12.9] (role), and [0.0, 3.5] (PoT), confirming severe degradation (p = 0.0005 for direct versus PoT).
For vision language gap, text only and multimodal confidence intervals do not overlap for closed-source models. GPT-5-Mini shows [82.9, 97.6] versus [62.4, 81.2], Claude [70.6, 89.4] versus [51.8, 72.9], and Gemini [64.7, 83.5] versus [51.8, 72.9] under direct prompting. Cohen's d effect sizes are large for both PoT degradation (d = 0.82 for GPT-5-Mini, d = 0.91 for Claude) and vision language gap (d = 1.24 for GPT-5-Mini, d = 1.08 for Claude), confirming practical significance.

\subsection{Performance Across Problem Families}

Table~\ref{tab:category_results} presents performance across 17 categories under direct prompting with multimodal inputs. Geometric and combinatorial problems achieve the highest success. GPT-5-Mini reaches 100\% execution on the traveling salesman, knapsack, k center, minimum spanning tree, capacitated flow, and transportation problems, with 0.0\% gaps, except for capacitated flow. Claude and Gemini show similar patterns. Llama-3.2-11B achieves at most 40\% execution with 97\%+ gaps. Flow conservation problems present challenges. Circulation achieves 60\% to 100\% execution but all models produce 100\% median gaps, indicating systematic constraint omission. Max flow shows gaps from 40.8\% (Gemini) to 141.1\% (Llama). Multi commodity flow combines execution and optimality challenges (GPT-5-Mini: 60\% execution, 14.1\% gap). Assignment and transportation show high variance. Transportation achieves 80\% to 100\% execution with 0.0\% gaps for closed source models. Assignment shows degraded performance (GPT-5-Mini: 80\% execution, 46.7\% gap). Transshipment achieves 0\% to 60\% execution. Bounded flow problems consistently underperform. Lower and upper bound flow achieves 0\% to 20\% execution. Network simplex achieves 0\% to 40\% execution, though Llama reaches 20\% with 1.8\% gap, possibly from memorization.

\begin{table*}[t]
\tiny
\centering
\caption{Performance by problem family under direct prompting with multimodal inputs. Dash = no successful executions.}
\label{tab:category_results}
\resizebox{\textwidth}{!}{%
\begin{tabular}{lcccccccccccc}
\toprule
& \multicolumn{3}{c}{\textbf{GPT-5-Mini}} & \multicolumn{3}{c}{\textbf{Claude-Haiku-4.5}} & \multicolumn{3}{c}{\textbf{Gemini-2.5-Flash}} & \multicolumn{3}{c}{\textbf{Llama-3.2-11B}} \\
\cmidrule(lr){2-4} \cmidrule(lr){5-7} \cmidrule(lr){8-10} \cmidrule(lr){11-13}
\textbf{Category} & \textbf{ESR} & \textbf{Gap} & \textbf{EMR} & \textbf{ESR} & \textbf{Gap} & \textbf{EMR} & \textbf{ESR} & \textbf{Gap} & \textbf{EMR} & \textbf{ESR} & \textbf{Gap} & \textbf{EMR} \\
\midrule
TSP & 100.0 & 0.0 & 100.0 & 100.0 & 0.0 & 100.0 & 80.0 & 0.0 & 80.0 & 0.0 & - & 0.0 \\
Knapsack & 100.0 & 0.0 & 100.0 & 100.0 & 0.0 & 80.0 & 100.0 & 0.0 & 100.0 & 20.0 & 546.2 & 0.0 \\
K Center & 100.0 & 0.0 & 100.0 & 100.0 & 0.0 & 100.0 & 100.0 & 0.0 & 100.0 & 20.0 & 100.0 & 0.0 \\
MST & 100.0 & 0.0 & 100.0 & 80.0 & 0.0 & 60.0 & 80.0 & 0.0 & 80.0 & 40.0 & 97.9 & 0.0 \\
Min Latency & 60.0 & 0.0 & 60.0 & 20.0 & 0.0 & 20.0 & 80.0 & 0.0 & 80.0 & 20.0 & 100.0 & 0.0 \\
\midrule
Circulation & 60.0 & 100.0 & 0.0 & 100.0 & 100.0 & 0.0 & 100.0 & 100.0 & 0.0 & 20.0 & 100.0 & 0.0 \\
Max Flow & 80.0 & 69.6 & 20.0 & 100.0 & 100.0 & 20.0 & 80.0 & 40.8 & 20.0 & 80.0 & 141.1 & 0.0 \\
Min Cost Flow & 60.0 & 26.4 & 0.0 & 20.0 & 16.4 & 0.0 & 40.0 & 0.0 & 40.0 & 0.0 & - & 0.0 \\
Multi Commodity & 60.0 & 14.1 & 20.0 & 40.0 & 2.8 & 20.0 & 20.0 & 0.0 & 20.0 & 40.0 & 100.0 & 0.0 \\
Capacitated Flow & 100.0 & 35.5 & 40.0 & 80.0 & 38.5 & 20.0 & 40.0 & 20.4 & 20.0 & 20.0 & 71.8 & 0.0 \\
Single Commodity & 80.0 & 41.5 & 20.0 & 60.0 & 27.2 & 20.0 & 100.0 & 37.0 & 40.0 & 0.0 & - & 0.0 \\
\midrule
Assignment & 80.0 & 46.7 & 20.0 & 20.0 & 29.0 & 0.0 & 20.0 & 29.0 & 0.0 & 0.0 & - & 0.0 \\
Transportation & 100.0 & 0.0 & 80.0 & 80.0 & 0.0 & 60.0 & 100.0 & 0.0 & 80.0 & 0.0 & - & 0.0 \\
Transshipment & 20.0 & 1.2 & 20.0 & 60.0 & 21.6 & 0.0 & 0.0 & - & 0.0 & 0.0 & - & 0.0 \\
\midrule
Lower/Upper Bounds & 20.0 & 0.0 & 20.0 & 0.0 & - & 0.0 & 20.0 & 0.0 & 20.0 & 0.0 & - & 0.0 \\
Network Simplex & 40.0 & 0.9 & 40.0 & 0.0 & - & 0.0 & 20.0 & 32.6 & 0.0 & 20.0 & 1.8 & 20.0 \\
Shortest Path & 60.0 & 7.1 & 20.0 & 100.0 & 22.2 & 0.0 & 80.0 & 9.1 & 20.0 & 20.0 & 100.0 & 0.0 \\
\bottomrule
\end{tabular}%
}
\end{table*}

\subsection{Error Analysis}

Manual analysis of 50 failures identifies three main categories. Visual parameter extraction errors (42\%) include misread capacities (18\%), wrong node labels (12\%), and missed edges (12\%). Constraint formulation errors (31\%) involve missing flow conservation (15\%) and incorrect bounds (10\%). Syntax and API errors (18\%) come from improper Gurobi usage or variable scope, while 9\% reflect incorrect objective direction. On text only inputs, visual errors vanish but constraint errors increase to 48\% and syntax errors to 35\%, confirming that visual extraction is the main multimodal bottleneck.

\subsection{Closed-Source Versus Open-Source Models}
On multimodal inputs under direct prompting, GPT-5-Mini achieves 71.76\% ESR versus 17.65\% for Llama-3.2-11B-Vision ($\Delta$ 54.11 pp). This persists on text-only inputs (92.94\% versus 16.47\%), indicating disparity beyond vision capabilities. EMR shows larger gaps: 44.71\% versus 1.18\% on multimodal inputs, confirming Llama rarely produces optimal solutions.

\section{Case Study: Deployment with MCP}
\label{sec:case_study}
\begin{figure}[t]
    \centering
    \includegraphics[width=0.8\columnwidth]{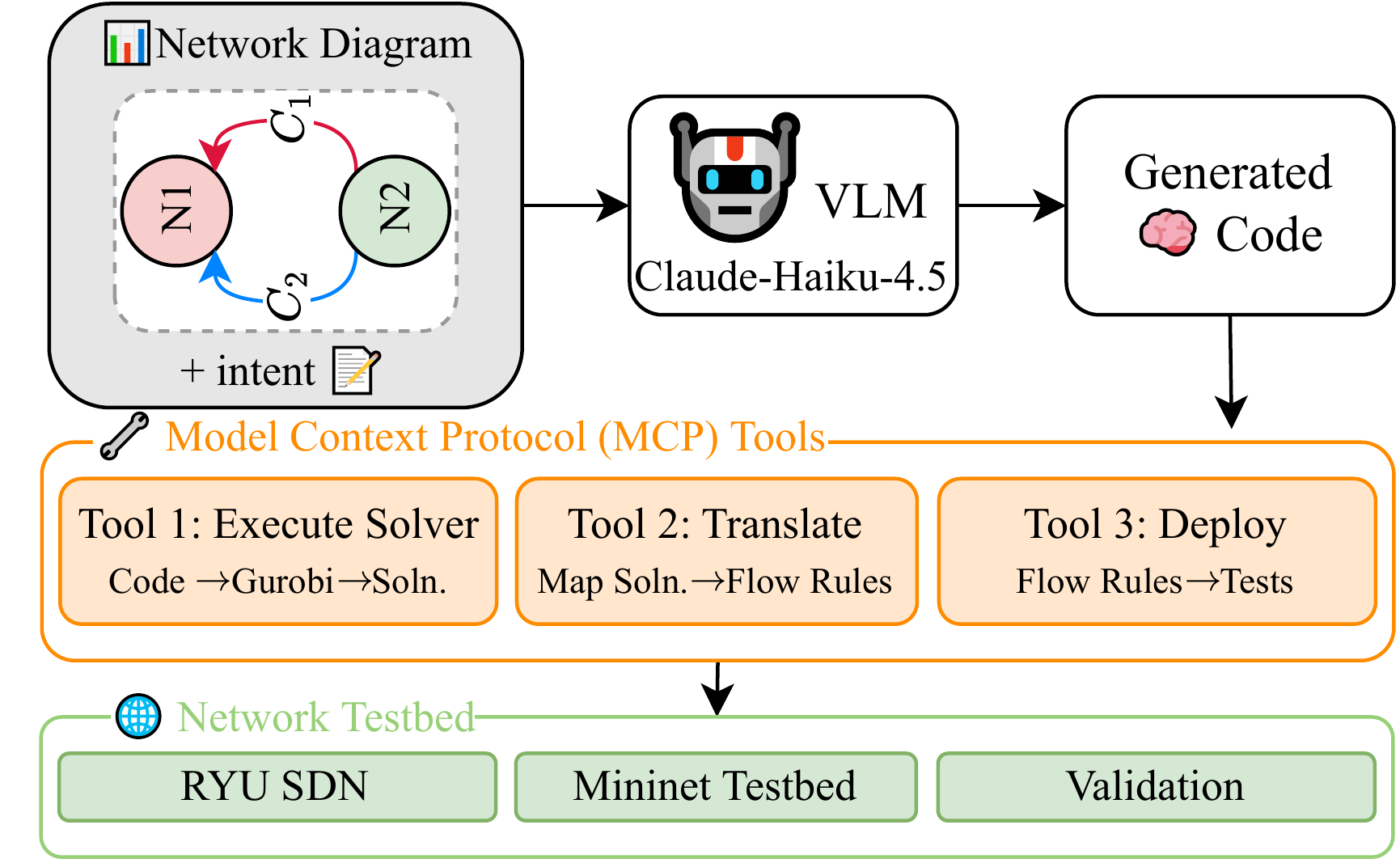}
    \caption{MCP deployment pipeline where the VLM generates optimization code from network intent, executes it, translates outputs to OpenFlow rules, and deploys them on the SDN.}
    \label{fig:mcp_pipeline}
\end{figure}
To evaluate practical feasibility, we implement a deployment pipeline that translates VLM-generated optimization solutions into network configurations and deploys them to a Mininet testbed with Ryu SDN controller (refer to figure~\ref{fig:mcp_pipeline}). The architecture integrates three MCP tools: \texttt{execute\_solver} runs generated code and extracts decision variables, \texttt{translate\_to\_openflow} converts solutions to OpenFlow rules, and \texttt{deploy\_and\_validate} installs rules via Ryu REST API and executes validation tests. The testbed comprises two hosts connected through an OpenFlow switch managed by Ryu controller.
We demonstrate the system on a minimum cost flow instance from our benchmark: two nodes with supply 193 and demand 193, connected by two links with capacities 99 and 199 and costs 2 and 7. Claude-Haiku-4.5 generates the Gurobi code that formulates and solves the problem, matching ground truth optimal value. Solver execution completes in 0.069 seconds. Translation produces three OpenFlow rules (bidirectional IPv4 forwarding and ARP handling) in under 1 millisecond. Rule installation via Ryu takes 1.8 seconds. Validation executes three tests (ping connectivity, capacity constraints via flow statistics, rule count verification), all passing in 2.1 seconds with zero packet loss. Total end-to-end latency is 12.2 seconds, with VLM code generation (8.2 seconds). The MCP architecture provides clean separation between optimization and network control, which allows deployment across multiple SDN controllers through unified tool interfaces.

\section{Limitations}

Our benchmark targets classical single objective optimization with deterministic parameters. Real network management often involves stochastic demands, uncertainty, and multi objective tradeoffs, which future extensions of IntentOpt could address. Evaluation is limited to Gurobi code generation. Testing across solvers such as CPLEX, Xpress, and OR Tools would assess portability. Current error analysis is based on 50 manually reviewed cases, and automated large scale classification could reveal more detailed patterns.
Code generation is evaluated in isolation, and full integration with IBN pipelines, including deployment and runtime monitoring, remains future work. The benchmark includes 85 instances across 17 families, and scaling to larger datasets would enhance statistical robustness. Visual diagrams follow standardized layouts, whereas real operator sketches are more varied.

\section{Conclusion}
We present IntentOpt, a benchmark of 85 network optimization problems spanning 17 categories, to evaluate VLM capabilities on multimodal optimization code generation for IBN. Our evaluation reveals three findings. First, visual parameter extraction introduces a 12 to 21 pp performance gap, with GPT-5-Mini dropping from 92.94\% to 71.76\% execution success. Second, PoT prompting degrades performance by up to 13 pp, suggesting verbose reasoning interferes with code synthesis. Third, open-source models substantially lag closed-source alternatives, with Llama-3.2-11B-Vision achieving 17.65\% execution success versus 75.29\% for GPT-5-Mini. These results establish current capabilities and limitations of VLMs for optimization intent processing, identifying clear directions for future research in vision encoders, domain adaptive pretraining, and open-source model development.

\bibliographystyle{IEEEtran}
\bibliography{ref}
\end{document}